%% file: main.tex
\documentclass[letterpaper, 10pt, conference]{IEEEtran}
\usepackage{times}
\usepackage{xr}
\usepackage{amsmath,amssymb,mathrsfs}
\usepackage[]{algorithm2e}
\usepackage{graphicx}
\usepackage[table]{xcolor}
\usepackage{wrapfig}
\usepackage{textcomp}
\usepackage{booktabs}
\usepackage{multirow}
\usepackage{textcomp}
\usepackage{textcomp} 

\definecolor{control}{RGB}{203, 65, 107}
\definecolor{shape}{RGB}{61, 153, 115}

\usepackage[numbers,sort]{natbib}
\usepackage{multicol}
\usepackage[bookmarks=true, hidelinks, colorlinks = true, linkcolor = black, urlcolor  = blue, citecolor = black, linkcolor=black]{hyperref}

\begin{document}

\title{Scale invariant robot behavior with fractals}

\author{Author Names Omitted for Anonymous Review. Paper-ID [add your ID here]}


\author{
\IEEEauthorblockN{Sam Kriegman$^1$,\,
Amir Mohammadi Nasab$^2$,\,
Douglas Blackiston$^3$,\,
Hannah Steele$^2$,\\[2pt]
Michael Levin$^{3}$,\,
Rebecca Kramer-Bottiglio$^2$,\,
Josh Bongard$^1$
}
\vspace{2pt}
\IEEEauthorblockA{%
$^1$University of Vermont, $^2$Yale University, $^3$Tufts University
}
}

\input{0_fig_teaser.tex}

\maketitle

\begin{abstract}
\input{0_abstract}
\end{abstract}

\IEEEpeerreviewmaketitle

\input{1_intro}

\input{0_fig_evolved_single}

\input{0_fig_evolved_designs}

\input{0_fig_manufacture}

\input{0_fig_fitness_curves}
\input{0_fig_hausdorff}

\input{2_methods}

\input{0_fig_biobots}

\input{3_results}

\input{4_discussion}

\input{5_thanks.tex}

\bibliographystyle{plainnat}
\bibliography{main}

\input{0_fig_lego_instructions}

\end{document}

%% file: 0_fig_teaser.tex
\teaser{
\centering
\vspace{-2pt}
    \includegraphics[width=\linewidth]{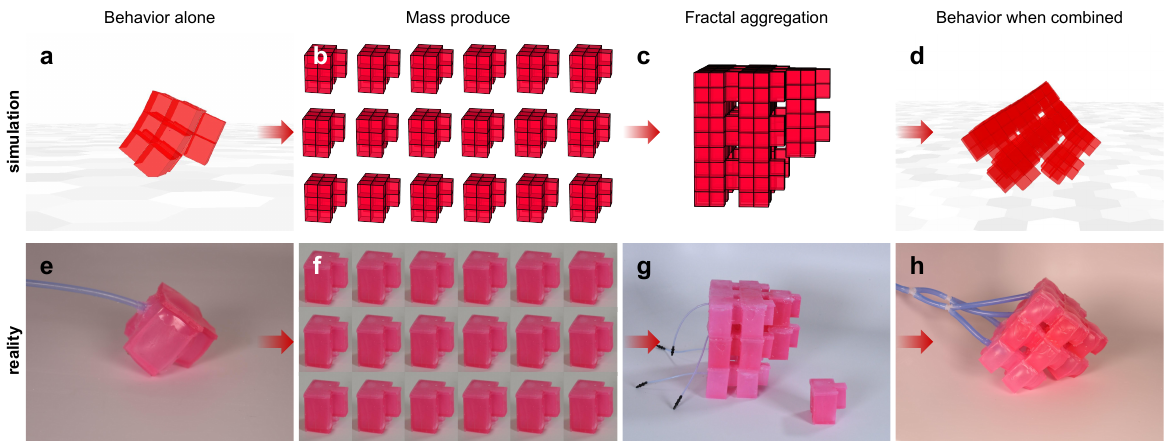}
\vspace{-18pt}
\caption{%
An evolved robot generates forward movement alone (\textbf{a},\textbf{e}), 
and then attaches with clones (\textbf{b},\textbf{f}) to form a fractal robot (\textbf{c},\textbf{g}) with the same behavior at a larger size scale (\textbf{d},\textbf{h}). 
More information, videos and code can be found at 
\href{https://fractalrobots.github.io}{\textbf{fractalrobots.github.io}}.
} 
\label{fig:teaser}
\vspace{-24pt}
}

%% file: 0_abstract.tex
Robots deployed at orders of magnitude different size scales, and that  retain the same desired behavior at any of those scales, 
would greatly expand the environments in which the robots could operate.
However it is currently not known whether such robots exist, and, if they do, how to design them.
Since self similar structures in nature often exhibit self similar behavior at different scales, we hypothesize that there may exist robot designs that have the same property. Here we demonstrate that 
this is indeed the case for some, but not all, modular soft robots: 
there are robot designs that exhibit a desired behavior at a small size scale, and if copies of that robot are attached together to realize the same design at higher scales, those larger robots exhibit similar behavior. We show how to find such designs in simulation using an evolutionary algorithm. 
Further, when fractal attachment is not assumed and attachment geometries must thus be evolved along with the design of the base robot unit, scale invariant behavior is not achieved, demonstrating that structural self similarity, when combined with appropriate designs, is a useful path to realizing scale invariant robot behavior. We validate our findings by demonstrating successful transferal of self similar structure and behavior to pneumatically-controlled soft robots. Finally, we show that biobots can spontaneously exhibit self similar attachment geometries, thereby suggesting that self similar behavior via self similar structure may be realizable across a wide range of robot platforms in future.

%% file: 1_intro.tex
\section{Introduction}
\label{sec:intro}

Fractals are ubiquitous in nature 
and increasingly prevalent in artificial structures \cite{puente1996fractal,fan2014fractal}.
Coastlines, rivers, and trees all exhibit self-similar structures---accurate replicas of themselves within themselves.
Fractals are also prevalent across length scales in animals and cellular structures:
respiratory and vascular systems, brains \cite{kiselev2003brain}, and DNA \cite{lieberman2009comprehensive}
all exploit 
space filling fractal networks.
At the largest known length scale, the universe itself consists of a fractal clustering of spiraling galaxies \cite{wu1999large}.
However, the ways in which fractals may be useful in robotics is mostly unexplored.

One desirable property of fractals derives from their ability to represent infinite complexity within a compact representation: a simple rule applied to itself, recursively.
\citet{hornby2001evolving} exploited this phenomenon in robotics by evolving branching kinematic chains in simulation and then building physical instances of the most promising designs \cite{hornby2003generative}.
The resulting robots yielded an order of magnitude more physical elements than any other simulated robots at that time:
Whereas \citet{sims1994evolving} and \citet{bongard2001repeated}
evolved simulated robots composed of no more than 14 and 50 parts,
respectively,
\citet{hornby2001evolving} utilized Lindenmayer-systems \cite{lindenmayer1968mathematical} to produce branching structures of up to 350 parts.

The ubiquity of self-similar forms in living systems suggests that fractals may confer adaptive benefits
in addition to reduced descriptive complexity, which could be useful in artificial systems such as robots. 
One such potential use of fractals in robotics is that self similar structure can, in some cases, result in self similar behavior \cite{khaluf2017scale}. Branching arteries and capillaries maximize efficient blood flow at the macro and micro scales, respectively \cite{mandelbrot1982fractal}. 
In the gecko, 
nanoscale spatulae branch from microscale setae, which in turn branch from toes, which in turn branch from feet, and together maximize the probability of adhesion 
of spatulae, setae, toes, and ultimately the animal itself to vertical surfaces \cite{autumn2000adhesive}.
However, with the exception of artificial gecko feet \cite{yu2018design} and Moravec and Easudes' hypothetical trillion-fingered Bush Robot \cite{moravec1999fractal}, this particular property of fractals remains unexplored in robotics. Here, we explore whether self similar structure can confer self similar behavior in modular robots.

Conventional robots contain numerous smaller components (legs, wheels, end effectors, servomotors, sensors, battery packs) but these parts are highly specialized, interdependent, and incapable of independent or self-similar behavior. 
Modular robots differ in that they consist of repeated robotic elements that are to some degree self-sufficient, able to behave and survive on their own.
They often autonomously detach, move about each other, and reattach to reconfigure the robot's overall geometry
\cite{romanishin20153d,jing2016end,pathak2019learning}.
But robot modules tend to be 
cubes or other simple geometries that do not reprise the shape of the whole robot, unless the robot's overall structure is 
also a simple geometry (e.g.~a cube of cubes).
In all cases, the target behavior of a module is, by design, different from that of the whole robot.

\input{0_fig_sponge}

Swarms of hundreds or thousands of individual robots have loosely coupled to form amorphous phototatic aggregates \cite{li2019particle} 
or moved within a shared shell \cite{savoie2019robot}, 
however these robots have largely been restricted to cylindrical geometries and 2D interaction in-plane \cite{rubenstein2014programmable}.
Most robot swarms comprise rigid-bodied electromechanical robots which, in order to be produced in large quantities, are designed to be as simple as possible.
A consequence of this is that each robot may be more or less behaviorally static (functionless)
in isolation
\cite{li2019particle,savoie2019robot}.

Here, we test whether self-similar forms can facilitate the evolution of 
scale invariant behavior as follows.
Shapes of basal robots (e.g.~Fig \ref{fig:teaser}a) are evolved in a voxel-based simulation; each design can be composed with copies of itself, using the same pattern,  indefinitely to realize self-similar form at a range of size scales.
However, if the desired behavior is lost and needs to be relearned \textit{de novo} at each level of recursion, then training such a system to operate on more than one size scale becomes computationally infeasible.
Thus, an evolutionary algorithm is here employed to find behaviorally scale-invariant fractal robots:
they must demonstrate similar behavior, normalized for scale, at three different size scales.
The results demonstrate that robots with this property can be automatically designed and 
in some cases, manufactured in reality.

In the first attempts to transfer simulated scale invariant behavior to a physical system,
computer-generated
mold designs were 3D printed and then used to rapidly fabricate identical, 
hollow elastomeric modules that can be  de/pressurized to induce volumetric actuation~\cite{kriegman2020sim2real4designs}.
Modules behaved in isolation and together when attached to form a fractal soft robot.

Scale invariant fractal design principles could also provide novel solutions to challenges facing \textit{in vitro} bioengineering.
One such challenge is the size limit of biobots without vascular systems in which diffusion does not bring sufficient oxygen and other key metabolites beyond approximately 1~mm.
As proof of principle, we show that ``xenobots'' 
(motile biological machines built from amphibian stem cells \cite{kriegman2020xenobots,blackiston2021new})
can be mechanically joined forming permanent multi-individual subunits that do not need vacular systems because they always keep their cells close to an interface with a nutrient medium.
These subunits can then be stacked and multiplexed to form increasingly large fractal biobots.

%% file: 0_fig_sponge.tex
\begin{figure}[t]
    \centering
    \includegraphics[trim={300pt 200pt 250pt 100pt},clip,width=\linewidth]{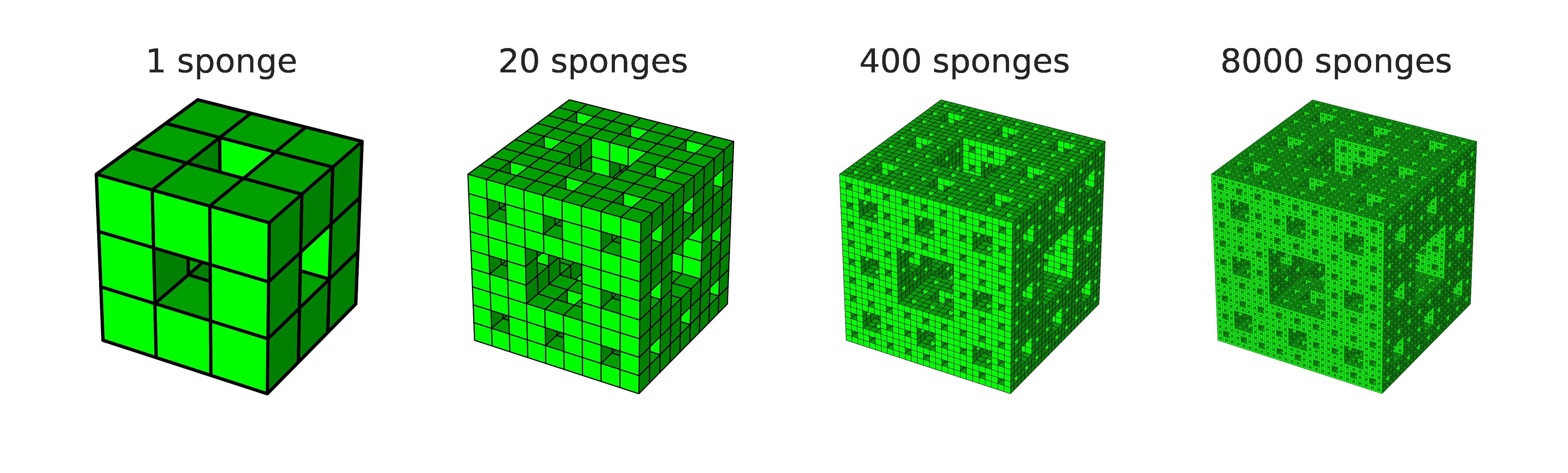}
    \vspace{-18pt}
    \caption{\textbf{Fractalizing mass-produced robots.} 
    As an example of how basal robots are here combined in fractals, the well-known Menger sponge is shown in place of an evolved design (1 sponge) at three fractal scales.
    A single (basal) sponge is a 3-by-3-by-3 cube of voxels with the middle voxel missing on each face, and no center voxel (a total of 20 voxels).
    Replacing each voxel with a copy of the entire sponge creates a fractal: a self-similar aggregation of 20 sponges.
    Repeating this operation yields larger and larger fractals.
    }
    \label{fig:sponge}
    \vspace{-12pt}
\end{figure}

%% file: 0_fig_evolved_single.tex
\begin{figure}[t]
    \centering
    \includegraphics[width=\linewidth]{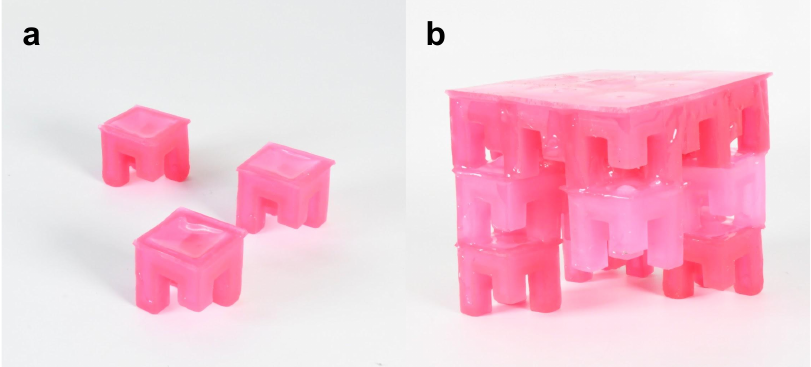}
    \vspace{-15pt}
    \caption{\textbf{Evolved single-actuator fractal.} 
    An asymmetrical quadrupedal fractal was the best fractal found by evolution when every basal robot (\textbf{a}) must synchronize their volumetric actuation in time.
    The entire structure (\textbf{b}) can thus be treated as a single bladder and pressurized by a single air inlet, at any size scale.
    Videos: the \href{https://youtu.be/FnHZBylM_JU}{simulation};
    the physical robot at the \href{https://youtu.be/u7jEvPp-CrU}{basal} and \href{https://youtu.be/jr2Toz6YCn8}{middle} scales. 
    At atmospheric pressure, the basal robot fits inside a 2$\times$2$\times$2 cm cube; its chamber walls are 2 mm thick. 
    }
    \label{fig:design2}
    \vspace{-12pt}
\end{figure}

%% file: 0_fig_evolved_designs.tex
\begin{figure*}[t]
    \centering
    \includegraphics[width=\linewidth]{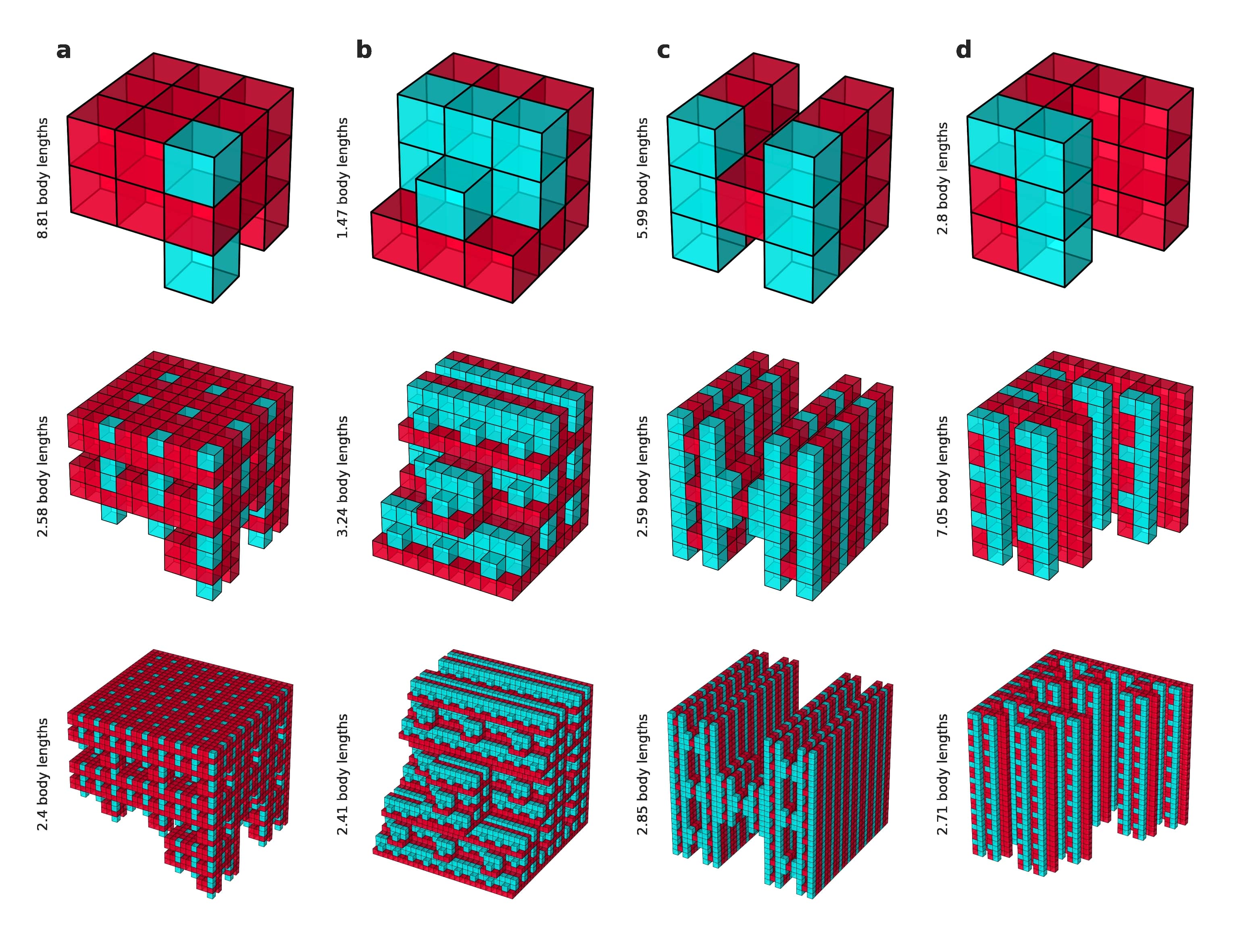}
    \vspace{-30pt}
    \caption{\textbf{Evolved fractal robots.}
    Four designs (columns; \textbf{a}-\textbf{b}) evolved for scale invariant locomotion in fractal aggregations at three size scales (rows): one robot (top; 3 cm wide), robots inside of a robot (middle; 9 cm), and robots inside of robots inside of a robot (bottom; 27 cm) (\href{https://youtu.be/EhQ4RakFRok}{video}).
    Voxels are 1 cm in length in simulation.
    Cyan and red voxels actuate in antiphase at 5 Hz.
    The fitness of a scale invariant design is the least amount of net displacement it generates across the three size scales (body lengths per minute; Eq.~\ref{eq:fitness}).
    }
    \label{fig:evolved_designs}
    \vspace{-12pt}
\end{figure*}

%% file: 0_fig_manufacture.tex
\begin{figure*}[t]
    \centering
    \includegraphics[width=\linewidth]{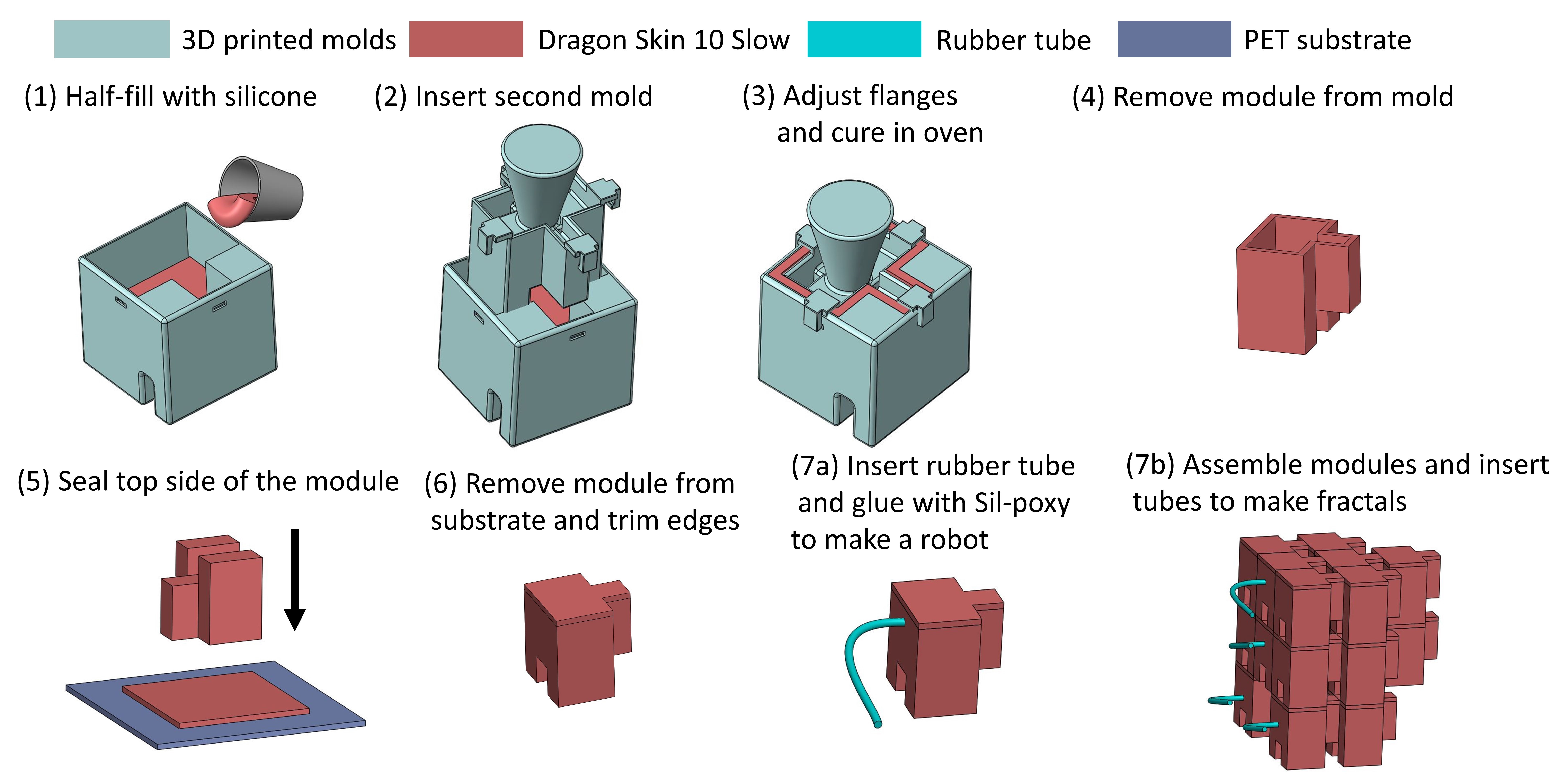}
    \vspace{-16pt}
    \caption{\textbf{Manufacturing fractal robots.}
    (1) 3D printed mold was partially filled with Dragon Skin 10 Slow and then placed in a vacuum oven to degas 
    for 5 minutes. (2) Second 3D printed mold was inserted into the first mold slowly to avoid introducing air bubbles into the silicone. (3) Flanges on the second mold were adjusted on the first mold to keep a 2 mm clearance between the walls of the molds. Silicone was cured in the oven at 60$^{\circ}$C for 75 minutes. (4) The module was removed from the mold and the edges were trimmed. (5) The module was flipped over an uncured layer of silicone and its top side was sealed. (6) The module was removed from the substrate and its edges were trimmed. (7a) A hole was punched in one of the sides of the module and a rubber tube was inserted and glued using Sil-poxy as the bonding agent to make a robot (\href{https://youtu.be/bHJR8srK8LI}{video}). (7b) Basal robots join together to form a fractal robot (\href{https://youtu.be/m6hYhY7g91E}{video}).
    }
    \label{fig:manufacture}
    \vspace{-1em}
\end{figure*}

%% file: 0_fig_fitness_curves.tex
\begin{figure*}[t]
    \centering
    \begin{minipage}{0.485\textwidth}
        \centering
        \includegraphics[width=\textwidth]{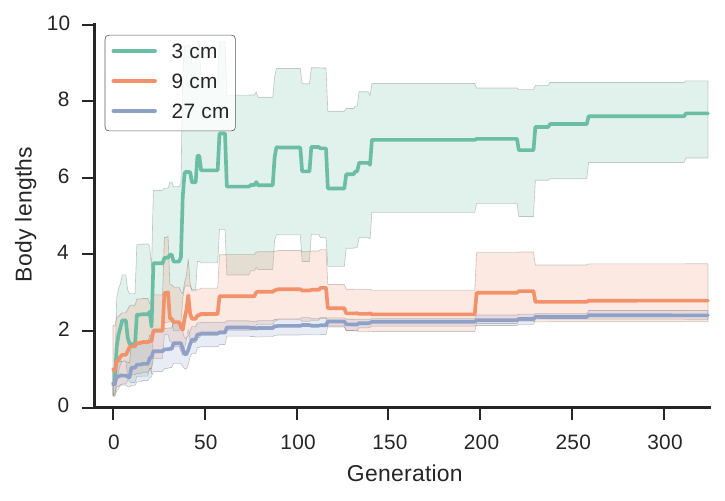} 
        \vspace{-2em}
        \caption{\textbf{Evolving scale-invariant behavior at three fractal scales.} 
        The mean fitness of the best design (in body lengths per minute) is plotted for ten independent evolutionary trials (95\% confidence intervals).
        Fitness was taken to be the worst performance out of the three size scales: 
        a single robot (3 cm), 
        a fractal aggregation of robots (9 cm),
        and a fractal aggregation of fractal aggregations of robots (27 cm).
        The null hypothesis is that of no evolutionary improvement in fitness (scale-invariant locomotion).
        On the basis of two samples (randomly-generated designs (generation 0) versus evolved designs (generation 325))---and a distribution-free rank sum test (Wilcoxon)---we reject the null hypothesis of no evolutionary effect ($p<0.001$).
        \label{fig:fitness_curves}
        }
    \end{minipage}\hfill
    \begin{minipage}{0.485\textwidth}
        \centering
        \includegraphics[width=\textwidth]{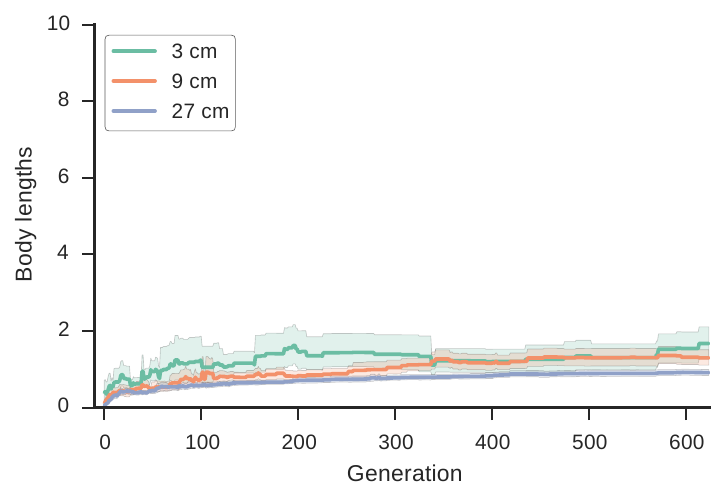}
        \vspace{-2em}
        \caption{\textbf{Evolving scale-invariant behavior without the fractal assumption.} 
        The mean fitness of the best design (in body lengths per minute) is plotted for 15 independent evolutionary trials (95\% CIs).
        Fitness is the worst performance out of the three fractal size scales (3 cm, 9 cm, 27 cm).
        The null hypothesis of interest is that there is no difference in fitness (scale-invariant locomotion) whether or not fractals are imposed.
        Thus, we have two samples: one that assumes fractals 
        (the performance curves in Fig.~\ref{fig:fitness_curves})
        and another that does not explicitly utilize fractals (the curves here in Fig.~\ref{fig:control}) (\href{https://youtu.be/oJsRpRkQ9F0}{video}).
        Using a distribution-free rank sum test (Wilcoxon), we reject the null hypothesis that there is no evolutionary effect of fractals ($p<0.001$).
        \label{fig:control}
        }
    \end{minipage}
    \vspace{-1em}
\end{figure*}

%% file: 0_fig_hausdorff.tex
\begin{figure}[t]
    \centering
    \includegraphics[width=\linewidth]{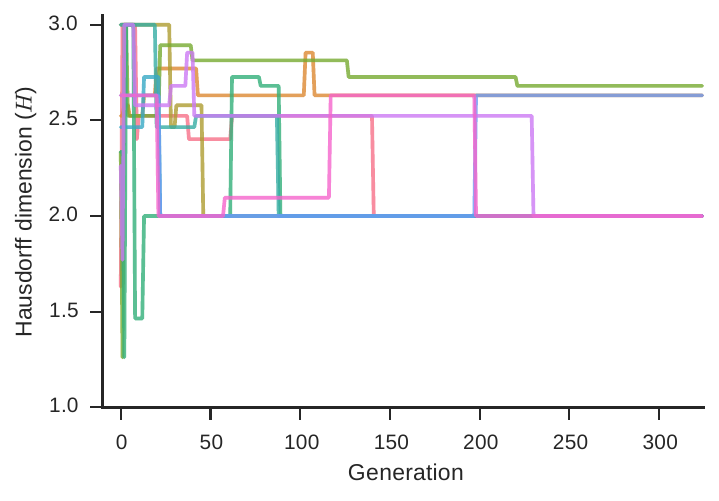}
    \vspace{-2em}
    \caption{\textbf{Hausdorff dimension across ten evolutionary trials.} 
    Smooth fractal geometries such as points, straight lines, flat planes, and cubes have Hausdorff dimension
    0, 1, 2 and 3, respectively.
    More rough fractal surfaces have non-integer Hausdorff dimension $(H)$.
    Each individual color plots $H$ of the most scale invariant design, at each generation, of a single evolutionary trial.
    Half of the trials ended in a plane of voxels ($H=2$), the the other half found rougher, more geometrically interesting designs with $2.5<H<2.7$.
    }
    \label{fig:hausdorff}
    \vspace{-1em}
\end{figure}

%% file: 2_methods.tex
\section{Methods}
\label{sec:methods}

\subsection{Fractalizing robots.}

Robots in this paper are modeled as polycubes with at most $c$ cubes (voxels) connected face-to-face inside a bounding box with length $m$ voxels.
A fractal aggregation is made by replacing each voxel in a basal robot with a copy of itself.
This operation can be repeated (to form an aggregate of aggregates of elements) either recursively---replacing voxels with the current aggregate---or iteratively---replacing voxels with the basal robot.
We use iteration here to achieve additional size scales with fewer elements.
However, the results of this paper are not reliant on the distinction between recursion and iteration.

The Menger sponge 
provides a familiar example (Fig~\ref{fig:sponge}).
A basal sponge is a box, $m=3$ voxels in length, with a hole through the center of each plane.
Replacing each of the $c=20$ voxels within an basal sponge generates a fractal sponge of 20 sponges.
Repeating this operation a second time generates a sponge of 20 sponges composed of 20 smaller sponges (400 sponges in total).
And so on.

The Menger sponge has
Hausdorff dimension:
\begin{equation}
    \label{eq:hausdorff}
    H= \log_m(c)= \frac{\log c}{\log m} = \frac{\log 20}{\log 3} \approx 2.727,
\end{equation}
which describes the space filled by a fractal geometry in the limit.
The Hausdorff dimension
is
zero for a single voxel;
one for a column of three voxels $(\log_3(3) = 1)$; 
two for a 3-by-3 plane of nine voxels $(\log_3(9) = 2)$;
and
three for a 3-by-3-by-3 cube of 27 voxels $(\log_3(27) = 3)$.
The design space here spans these (smooth) structures and many more in between.
Since all fractals here have $m=3$, $H \propto \log c$.

Hausdorff dimension provides an interpretable measure of the structural complexity (or ``roughness'') of the robots evolved in this paper.
It will allow us to analyze how fractals with scale invariant behavior can emerge from initially random fractals over evolutionary time.

\subsection{Simulating fractal robots.}

Voxels are simulated by a point mass and up to six Euler-Bernoulli beams connecting to neighbors on each voxel face.
There are two kinds of voxels (red and blue) which actuate in antiphase.
Volumetric actuation is implemented by treating the beams as pistons and adjusting their rest length at every time step according to a sine wave with a phase shift stored at each voxel.
A given voxelyzed robot can be fractalized by replacing each voxel with a copy of the entire robot, just like the Menger sponge.

To simulate the high mechanical resolution of fractal robots we used voxcraft-sim \cite{liu_voxcraft_2020},
a GPU-accelerated re-implementation of Voxelyze 
\cite{hiller2014dynamic}.
In Voxelyze, voxels are evaluated sequentially on a single thread of a CPU, one after another. 
In voxcraft-sim, 
voxels are evaluated concurrently on GPUs. 
Collisions in Voxelyze are resolved in a double loop of all $n$ surface voxels, with time complexity $\mathcal{O}(n^2)$.
In voxcraft-sim, collisions are handled using a bounding volume hierarchy tree 
\cite{karras2012maximizing} 
with $\mathcal{O}(n\log n)$.

The basic simulated elements of Voxelyze and voxcraft-sim are the same (beams and masses) but computing dynamics in parallel and handling collisions with a BVH tree makes it possible to simulate
geometrically interesting robots with complicated surface areas that self collide.
Here, evolution operates in a 27$^3$ voxel workspace (up to 19,683 total voxels).
Due to past computational limits, previous evolutionary robotics work was largely restricted to voxel workspaces with less than a thousand voxels.
The lone exception was
\citet{hiller2012automatic}, who
evolved 
soft robots
within a 20$^3$ voxel workspace (up to 8000 voxels). 
However, 
strong morphological 
convergence 
led to a scooting 
mass that filled less than a quarter of the workspace 
(less than 2000 voxels).
Here, we demonstrate how fractals can provide a more scalable encoding.

\subsection{Evolving fractal robots.}

\textit{Encoding.}
Each basal robot design is generated by an evolved Composition Pattern-Producing Network 
(CPPN; \citep{stanley2007compositional,cheney2013unshackling}).
Each network ``paints'' 
a pair of floating point values in the range [-1,1] at every (x,y,z) location in the 3-by-3-by-3 workspace.
The first value is thresholded at 0 to determine the presence or absence of material at that point.
The second is thresholded at 0 to determine the type of material (if any) at that point.
The largest connected component of material output by a network (genotype) is taken to be the morphology of the basal robot (phenotype).
For more details see \cite{kriegman2020xenobots}.

\textit{Fitness.}
Each basal robot design is evaluated three times:
first alone, and then twice more in fractal aggregations of increasing size.
Performance at the $i$-th size scale is measured by the net displacement $d_i$ the robot or aggregate traveled (measured in body lengths) at the end of the 5 second evaluation period (i.e.~25 actuation cycles at 5~Hz).
A design's fitness, $F$, is taken to be the least performance achieved across the three size scales:
\begin{equation}
    F = \min(d_1, d_2, d_3).
    \label{eq:fitness}
\end{equation}

\textit{Simulation hyperparameters.}
Actuation here occurs at 5~Hz, with amplitude $A=\pm50\%$ resting volume.
Material was simulated with 10 kg/m$^3$ density,
Young's modulus $10^4$ Pa, and Poisson's ratio 0.5.
Each voxel is 1 cm in length.
A Coulomb friction model is used for the surface plane.
Coefficients of static and dynamic friction were 1 and 0.5, respectively.
Beams and collisions are damped with $\zeta=1$ and $0.8$, respectively.

\textit{Algorithm hyperparameters.}
Populations of 
CPPNs were evolved with ``age protection''~\citep{schmidt2011age}, 
an additional selection pressure that maintains diversity by relaxing competition on newer genetic lineages.
A randomly-generated 
CPPN,
with age 0, is injected into the population at every generation.
Mutations add/remove/modify vertices and edges within a selected network.
The activation functions at each vertex are randomly chosen from the following  functions: sine, absolute value, square, square root, and a step function.
For more details see \cite{kriegman2020xenobots}.
The radial distance from the center of the workspace and a bias of 1 were also input to each network, in addition to cartesian coordinates x,y,z.

Source code can be found at 
\href{https://fractalrobots.github.io/code}{fractalrobots.github.io/code}.

\subsection{Manufacturing fractal robots.}

Previous work \cite{kriegman2019automated,kriegman2020sim2real4designs} used a rotational molding technique to fabricate hollow cubic silicone modules (physical voxels) as building blocks of pneumatic robots. 
Here we introduce a molding technique, which reduces the weight of silicone required for module fabrication, and allows for the creation of custom-shaped basal modules.
 
Evolved fractal robots were fabricated using Dragon Skin 10 Slow
in a multistep molding and assembly process (Fig.~\ref{fig:manufacture}). 
Two rigid molds for casting were 3D printed.
The outer mold was treated with a release agent (Ease Release 200), half-filled with silicone, and then degassed in a vacuum oven at room temperature for 5 min.
Next, the inner mold was treated with release agent, and inserted into the outer mold.
Four flanges on four sides of the inner mold ensure that it remains centered with a 2 mm clearance from the outer mold.
Silicone was then allowed to cure in an oven at 60$^{\circ}$C for 75 min. 

Once cured, the module was removed from the mold, flipped over, and placed on a draw-coated layer of uncured silicone with 2 mm thickness to seal its top side. 
A hole was poked into one of the side walls of the module and a rubber tube was inserted and glued with Sil-Poxy 
to create the basal robot. 

To assemble a fractal aggregation, single modules were attached to each other using silicone elastomer as bonding agent. 
Finally, holes were punched in the side walls of the adjacent modules located in each horizontal plane. This allows groups of modules within the same plane to be pressurized by a single air inlet. 
For example, the robot in Fig. 1g,h has two groups of modules in the lowest horizontal plane (two “feet”), one group of modules in the middle plane, and one group on the highest plane, requiring just four independent air inlets, instead of 18.
Once assembled, robots were placed on top of a sheet of ABS Plastic (MyStudio MS20CYC Background Cyclorama), which was covered in cornstarch (Argo\textregistered, ACH Food Companies, Inc.) to reduce the friction between the silicone body of the robot and the plastic substrate~\cite{kriegman2020sim2real4designs}.


\subsection{Culturing fractal biobots.}

Living biobots, called xenobots \cite{kriegman2020xenobots}, were produced using two methods.
In the first, the animal cap of \textit{Xenopus} embryos (Nieuwkoop and Faber stage 10) were removed using surgical forceps and cultured in a mild saline solution (0.75x Marc’s Modified Ringers [MMR], pH 7.8) for 24 hours at 14$^{\circ}$C.  
Individuals do not naturally adhere to each other but can be induced to do so via wounding the outer surface with forceps.
Following injury, individuals are manually held together for five seconds, after which they heal together permanently.

For the second method, animal cap tissue is harvested from \textit{Xenopus} embryos as above and moved to a calcium/magnesium free dissociation media.  
This solution allows the pigmented outer layer of cells to be separated and discarded, leaving the white inner layer which form spheroids when moved to a mild saline solution (0.75x Marc’s Modified Ringers [MMR], pH 7.8).  
After 24 hours of development at 14°C, these spheroids will naturally adhere to neighbors upon touching, allowing multi-individual chains to be formed through simple contact.

%% file: 0_fig_biobots.tex
\begin{figure*}[t]
    \centering
    \includegraphics[width=\linewidth]{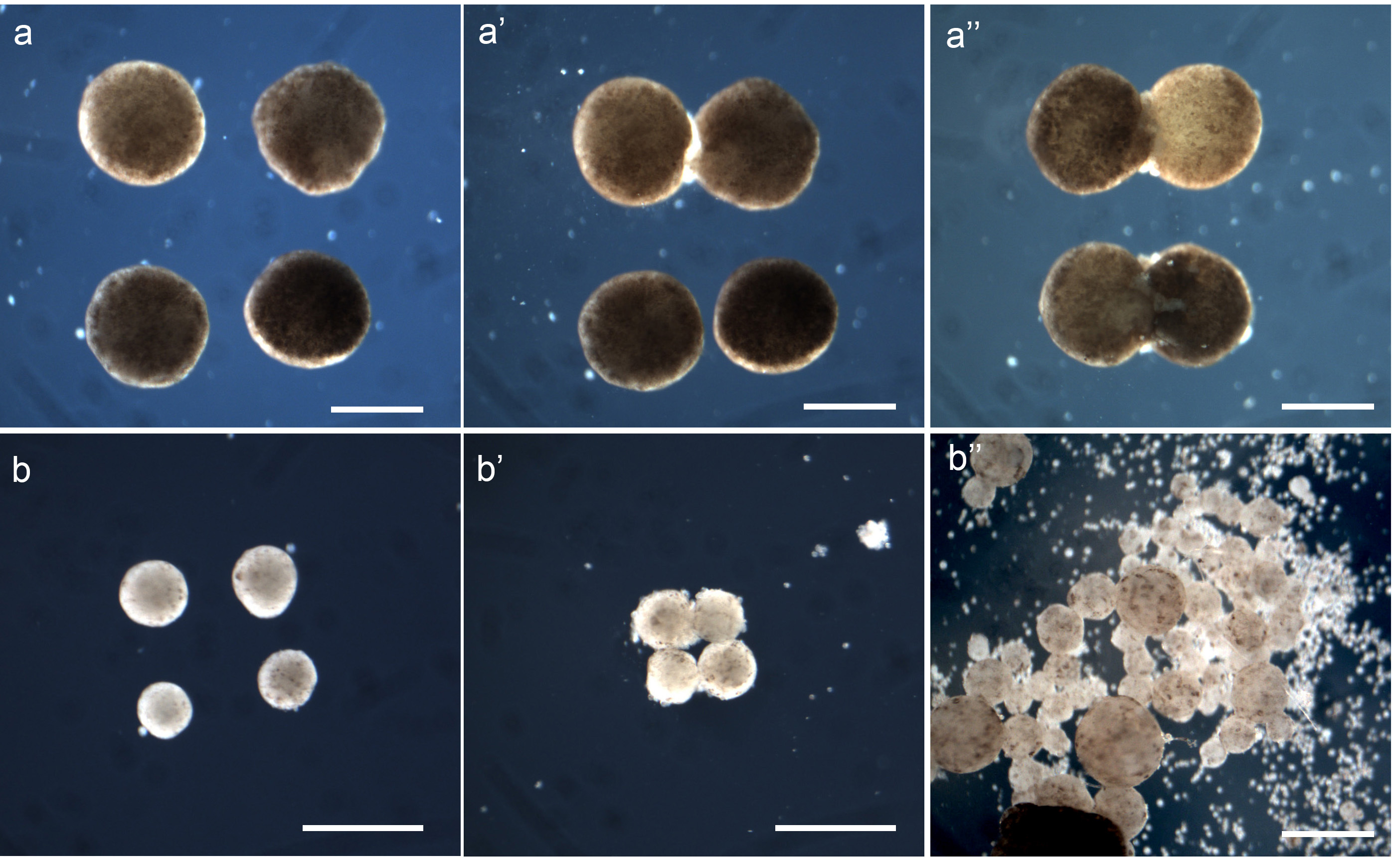}
    \vspace{-16pt}
    \caption{\textbf{Culturing fractal biobots.} Embryonic amphibian stem cells (\textit{Xenopus laevis} species) form spheres of motile (ciliated-driven) epidermis (a) without additional intervention.  
    Spheroids can be combined into various macro-combinations (\textbf{a'}, \textbf{a''}) through the induction of a laceration, which allows individual xenobots to fuse at the wound site (\href{https://youtu.be/qgsz8Sq621I}{video}).  
    Using a separate construction method which removes the outer pigmented cell layer (\textbf{b}), spheroids are naturally sticky and can be daisy-chained through simple contact (\textbf{b’}) enabling rapid connection of tens or hundreds of individuals (\textbf{b’’}) (\href{https://youtu.be/YCvmK_UBF9U}{video}). 
    Scale bars: 500µm.
    }
    \label{fig:biobots}
    \vspace{-1em}
\end{figure*}

%% file: 3_results.tex
\section{Results}
\label{sec:results}

\subsection{Simulated fractal robots.}

Ten independent evolutionary trials were conducted, each starting from a different set of random initial conditions.
Basal robots were evolved inside a
3-by-3-by-3 voxel workspace ($m=3$) using up to $c=27$ voxels, which are each 1 cm in length.
Basal robots were recursed twice to yield 
9-by-9-by-9 (up to 729 voxels) 
and
27-by-27-by-27 (up to 19,683 voxels) fractal robots.

Each trial used 8 NVIDIA Tesla V100s.
No trial took less than 40 hours or more than 43 hours to complete 325 generations of evolution.
Each generation consisted of 48 simulations (16 designs in the current population, each evaluated at three size scales), for a total of $325*48=15{,}600$ design evaluations per trial.

\textit{Scale invariant behavior.} Fig.~\ref{fig:fitness_curves} shows evolutionary improvement occurring on all three size scales.
Usually (but not always) the largest fractals (27 cm) generated less forward movement than the middle (9 cm) and basal (3 cm) scales, as evidenced by the low orange and blue curves in Fig,~\ref{fig:fitness_curves}.
There was not a significant difference in performance between the middle (9 cm) and largest (27 cm) size scales.
Because fitness is driven by the size scale with the worst performance (Eq.~\ref{eq:fitness}), after $\sim$50 generations of evolution, there seems to be, on average, little to no selection pressure on the performance of the basal and middle fractal scales.
This can be seen to some extent in the aggregate curves of Fig.~\ref{fig:fitness_curves}.
Whenever a one of the curves drops in performance over successive generations, the average improvement in scale invariant behavior was driven by another size scale.
The largest fractal scale more consistently trends upward in performance, relative to the middle and basal fractal scales.
This is likely due to the increased mass, but fixed elastic modulus and actuation amplitude, of the robot at larger fractal scales.

\textit{Evolved fractals.} 
The confidence bands visible in Fig.~\ref{fig:fitness_curves} indicate that
evolution did not converge to a single fractal design across the ten trials.
The diversity of form can be seen in the
four evolved robots drawn in 
Fig.~\ref{fig:evolved_designs} at the three fractal scales.
These four robots have interesting roughness:
fractal geometries that fill much more than a 2D plane but less than a full 3D cube.
Their roughness is formally measured by their Hausdorff dimension (Eq.~\ref{eq:hausdorff}), which is plotted over evolutionary time in
Fig.~\ref{fig:hausdorff}.
Early in evolution there were some full cubes $(H=3)$.
At the end of evolution there were five full planes $(H=2)$ and five designs with more roughness $(2.5<H<2.7)$.
As the basal shape changes in each column of Fig.~\ref{fig:evolved_designs}, so too does the distribution of performance at each size scale.
For some shapes, performance was highest at the smallest fractal scale (Fig.~\ref{fig:evolved_designs}a,c).
For others, performance was highest at the middle fractal scale (Fig.~\ref{fig:evolved_designs}b,d).
Some have more uniform performance across size scales (Fig.~\ref{fig:evolved_designs}b,c).
Others have one size scale in which performance is two or three times as high as their next best (Fig.~\ref{fig:evolved_designs}a,d).

\textit{Comparison to non-fractals.}
To determine how much (if at all) fractals aid the evolution of scale invariant behavior,
we compared against an otherwise equivalent algorithm that freely combines 3~cm basal robots into any aggregate structure within a 9~cm$^2$ bounding box, without imposing self-similarity.
At the third fractal size scale, aggregates are freely combined within a 27~cm$^2$ bounding box to form an aggregate of aggregates of basal robots.
Because we removed the assumption of self-similarity, some rule must be introduced to orchestrate aggregation at every size scale.
Because additional selection pressures are necessary to induce modularity in separate outputs of the same CPPN \cite{clune2010investigating},
we here use three independent CPPNs to dictate the shape at the three size scales.

Note that if modularity was not artificially induced, 
then the aggregation strategies produced by a single CPPN would by default be strongly correlated, yielding fractals, and thus defeating the purpose of this control experiment.
However, fractals can emerge spontaneously under these conditions if the independent CPPNs converge to identical aggregation functions that mirror the basal robot's internal voxel structure.


The behavior of a random nonfractal robot (before evolution) can be seen \href{https://youtu.be/oJsRpRkQ9F0}{here}.
The results after evolution indicate that searching for multiple aggregation policies, even just a two size scales, can significantly slow the evolution of scale invariant behavior compared to assuming fractality (Fig.~\ref{fig:control}).

\textit{Transferal to reality.}
Two additional evolutionary trials were conducted.
In the first, we used phase shifted actuation:
Instead of using cyan and red voxels that actuate in anti-phase, 
a wave of phase shifted actuation propagates through the body, from anterior to posterior, repeating every 0.2 seconds.
This experiment was conducted a few weeks before the experiments using anti-phase materials (Fig.~\ref{fig:evolved_designs}).
Thus, the robot with the most scale invariant behavior that evolved under these conditions (Fig.~\ref{fig:instructions_phase_offset}) was the first to be manufactured in reality (Fig.~\ref{fig:teaser}).

After observing the transferal of a phase shifted actuation,
a single bladder design was evolved in which all basal robots synchronized their actuation in time (zero phase-offsets across the body).
The evolved single actuator fractal robot with the most scale invariant behavior was then fabricated.
The winning design is shown in simulation and reality in Figs.~\ref{fig:design2} and \ref{fig:instructions_single_actuator}, respectively.

\subsection{Physical fractal robots.}

The best evolved fractal robot utilizing a phase shifted actuation (Fig.~\ref{fig:teaser}) was fabricated first (Fig.~\ref{fig:manufacture}). 
At the basal fractal scale, the robot was actuated with $\simeq$~6 kPa at a frequency of 3~Hz and a locomotion behavior was observed similar to the prediction from simulation. 
At the middle fractal scale, the robot was first actuated using a cam system which was powered by a DC motor (Pololu 131:1) with maximum rotational speed of 72 RPM and designed to operate two piston-cylinder assemblies (Clippard Inc.) for pressurizing/depressurizing the robot. 
This design led to a controlled actuation of the robot with maximum $\simeq$~4 kPa pressure at a maximum frequency of 1.2 Hz. This combination of the pressure and frequency did not lead to locomotion for the robot. 
Due to this limitation of the cam system, a secondary testing method was also employed in which the pressure inlet was manually switched between an airline with 20~kPa pressure and a vacuum pump for 50 cycles resulting in an average actuation frequency of 0.55 Hz. 
The robot was then able to locomote at the middle fractal scale, but at a much lower speed compared to the basal scale, due to hardware constraints in applying similar combination of input pressure and actuation frequency.

The best evolved single actuator fractal robot was the second design to be fabricated (Fig.~\ref{fig:design2}).
In the single actuator case, scale invariant behavior successfully transferred: the physical robot behaved similarly at the basal and middle fractal scales.
However, the direction of the robot's movement did not match the prediction from simulation.
We suspect that this is due to an overly simplistic (Coulomb) friction model. 
Given that the direction of locomotion for soft robots can be reversed by tuning the tribological properties of the surface plane \cite{majidi2013influence},
future work should either modify the simulator, 
or the physical surface on which the robot operates, or both.

\subsection{Living fractal biobots.}

In Fig.~\ref{fig:biobots} living xenobots \cite{kriegman2020xenobots} were formed from amphibian stem cells harvested from embryos of the African clawed frog \textit{Xenopus laevis}.  
The apical surface of individual spheroids are not naturally adherent to neighbors, however, following mechanical laceration with microsurgery forceps individuals can be attached to one another during the healing process.  
This attachment is stable (connected members can be pushed with forceps or moved via pipetting without detaching) and permanent, although members can be manually decoupled through bisection with forceps or a scalpel.

If the pigmented outer epidermal layer is removed during construction, the resultant biobot is naturally adherent to neighbors through contact---allowing for the rapid connection of many individuals.  
Long term stability is less using this construction method as the increased adherence can result in neighbors fusing into one large spheroid.  
Future work would allow these connected systems to be stacked vertically, creating three dimensional fractalized designs.

%% file: 4_discussion.tex
\section{Discussion}
\label{sec:discussion}

In this paper, scale invariant robot behavior was achieved using fractals.
The results demonstrate that, under certain conditions, the nonrandom behavior of a single robot can be preserved when connected with copies of itself fractally.
However, this depends on the morphology of the basal robot.
A majority of the fractals tested did not exhibit the same behavior at multiple size scales.
An evolutionary algorithm was used to weed out robots that behaved incorrectly at any of the tested size scales.

Fractal robots were fabricated in reality using two distinct hardware platforms: 
pneumatic soft robots and ciliated biobots.
Using pneumatically actuated silicone bladders, two centimeters across, 
self similar structure was readily assembled, and the physical robot exhibited scale invariant behavior.
However, the amount of locomotion was limited and the direction did not always match the simulated prediction.
The micrometer length scale biobots investigated here exhibited spontaneous attachment and scale invariant behavior up to supermillimeter length scales. However, self similar structure was more challenging to produce out of plane.

The following discussion highlights the key limitations of the current methods and important opportunities for future work.

\subsection{Limitations of current fractal design.}

For locomotion on a flat surface plane, larger robots can move farther in terms of raw distance but not necessarily in terms of body lengths (Fig.~\ref{fig:fitness_curves}), and they cannot fit into smaller 3D spatial constraints.
If a robot's task environment is constrained spatially in one dimension (e.g.~a low overhang) or two dimensions (e.g.~a narrow pipe) then the Mengerian fractalization (Figs.~\ref{fig:sponge} and \ref{fig:evolved_designs}) used here may not be appropriate.
Thus, instead of uniformly repeating basal robots in 3D space, future work will explore allometry: how fractals may expand at different rates along different axes, as dictated by physical constraints and selection for specific target behaviors.

It is also possible that the coordinate system itself was a poor choice.
A cubic grid was assumed because the simulator we used was voxel-based.
However, fractals formed using using spherical modules \cite{liang2020freebot,swissler2020fireant} and spherical coordinates (Mandelbulbs) may be better suited for forms and functions that depend on radial symmetries.


In addition to self-similar geometry and movement,
a system's statics (e.g.~structural integrity) could also be fratalic.
Such a robot may not look the same at all scales but it would resist forces passively in the same way.
Scale invariant statics could also provide an easier target than scale invariant movement for system identification and calibrating simulations prior to optimization and mass manufacture.


Finally, it should be noted that the choice of a synchronous evolutionary algorithm \cite{schmidt2011age} was somewhat arbitrary. 
It is likely that other computational search methods that are capable of producing self similar robot forms \cite{van2019spatial,zhao2020robogrammar} could be modified to generate scale invariant behavior in a similar fashion.

\subsection{Limitations of current physical fractal robots.}

As the number of modules within a fractal robot increases at higher size scales, so does the volume of the air required to attain a given pressure level.
This makes pressurizing a larger fractal robot at high frequency more challenging. 
The modules fabricated here (Fig.~\ref{fig:teaser}e) had a chamber volume of 18 cm$^3$. 
When 18 of these modules were assembled to form a fractal (Fig.~\ref{fig:manufacture}h) the chamber volume increased to $18^2=324\;$cm$^3$, which was more difficult to actuate at higher frequencies and thus slowed the robot's locomotion speed. 
To facilitate the actuation of the physical fractal robots, the volume of the base module should be miniaturized as much as the fabrication procedure allows. 
The number of pistons in the cam testing system could also be increased to allow for a higher pressure within robot while maintaining a high frequency.

The designs transferred from simulation to reality in this paper served as a proof of principle that fractal robots with scale invariant behavior could in fact be realized. 
However, it may be impossible in practice to retain a \textit{useful} behavior across a meaningful range of length scales using current soft robot technologies, such as those that rely on pneumatic actuation.

\subsection{Limitations of current fractal biobots.}


The construction of fractalized biobots faces a number of biological challenges and constraints.
First, the developmental timing permissive for attachment is narrow and requires frequent manual manipulation.
If development is too early (less than 24 hours of culturing), individuals fuse completely forming one large sphere, and if development is too late (48 hours or more) the outer surface becomes covered naturally occurring mucus, making attachment difficult.
Second, individuals must be manually connected and observed for healing for 10-20 minutes each, necessitating significant time for husbandry from the investigator.
Finally, living biological system are not static and continue to remodel throughout their lifespan.
Cell division and cell death are natural components of all developmental systems, which makes the long-term stability of such fractalized designs an interesting area of study.




%% file: 5_thanks.tex
\section*{Acknowledgements}

This work was supported by
DARPA contract HR0011-18-2-0022,
NSF EFRI award 1830870,
and
NSF NAIRI award 2020247.

%% file: 0_fig_lego_instructions.tex
\begin{figure*}[t]
    \centering
    \includegraphics[width=\linewidth]{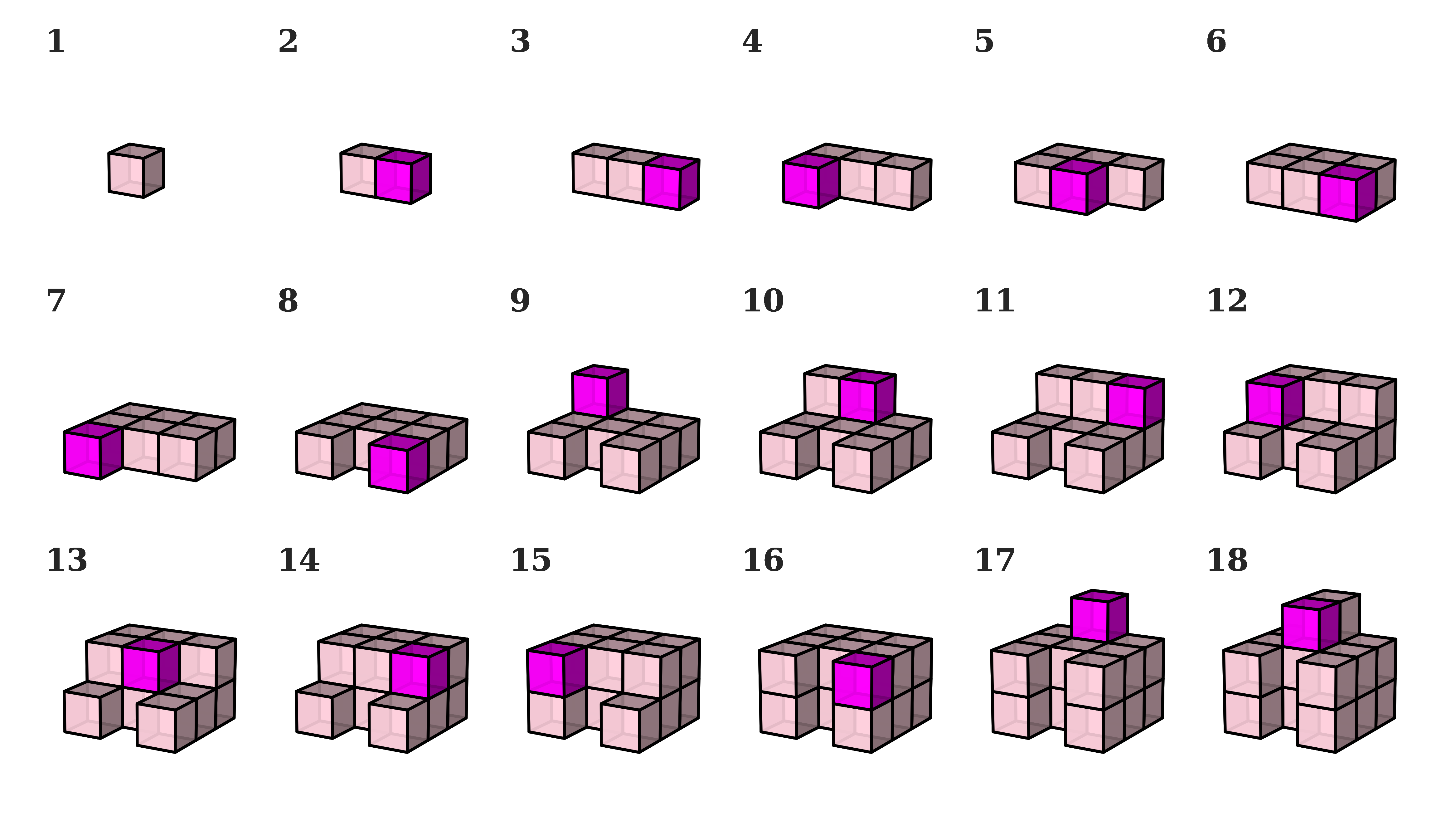}
    \includegraphics[width=\linewidth]{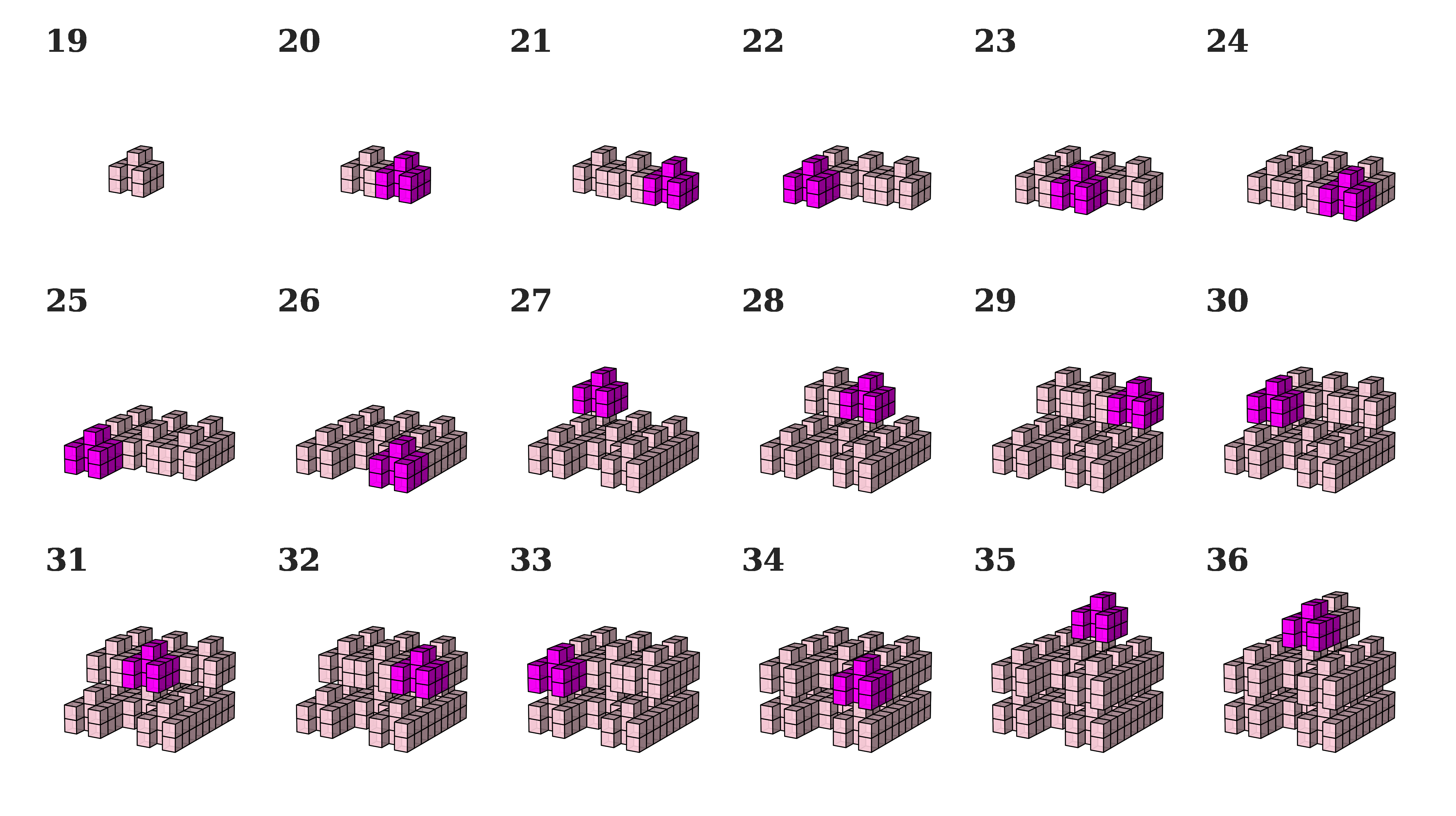}
    \caption{\textbf{Instructions for building the evolved fractal with the most scale invariant behavior using phase shifted actuation.}
    The local volume of the robot fluctuates according to a sine wave with a preselected phase offset at each module (dark pink):
    each wave of actuation propagates through the body, from anterior to posterior, repeating every 0.2 seconds 
    (\href{https://youtu.be/okLwbe17-9Y}{video}). 
    The basal robot (pictured in step 19) was designed in simulation by combining up to 27 voxels inside a 3-by-3-by-3 workspace (steps 1-18).
    The same steps were repeated using the basal robot instead of a single voxel (steps 19-36) to create a self similar aggregate machine.
    Note that the voxels only exist in simulation as design tools; there are no physical voxels; a custom mold is made according to the basal robot's shape (Fig.~\ref{fig:manufacture}).
    The physical robot assembled according to these instructions can be seen, turned right side up, in Fig.~\ref{fig:teaser}e-h.
    }
    \label{fig:instructions_phase_offset}
\end{figure*}

\begin{figure*}[t]
    \centering
    \includegraphics[width=\linewidth]{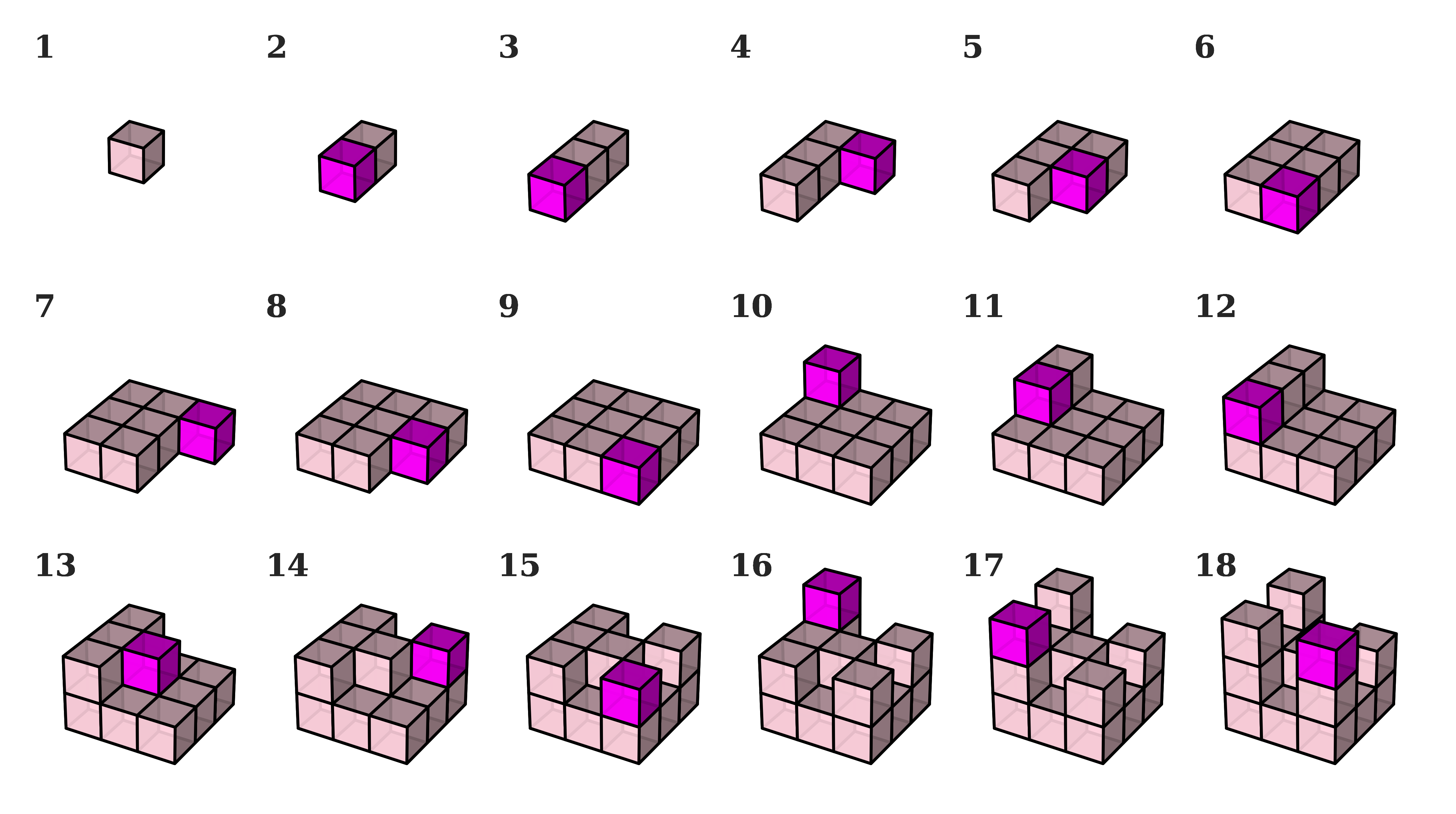}
    \includegraphics[width=\linewidth]{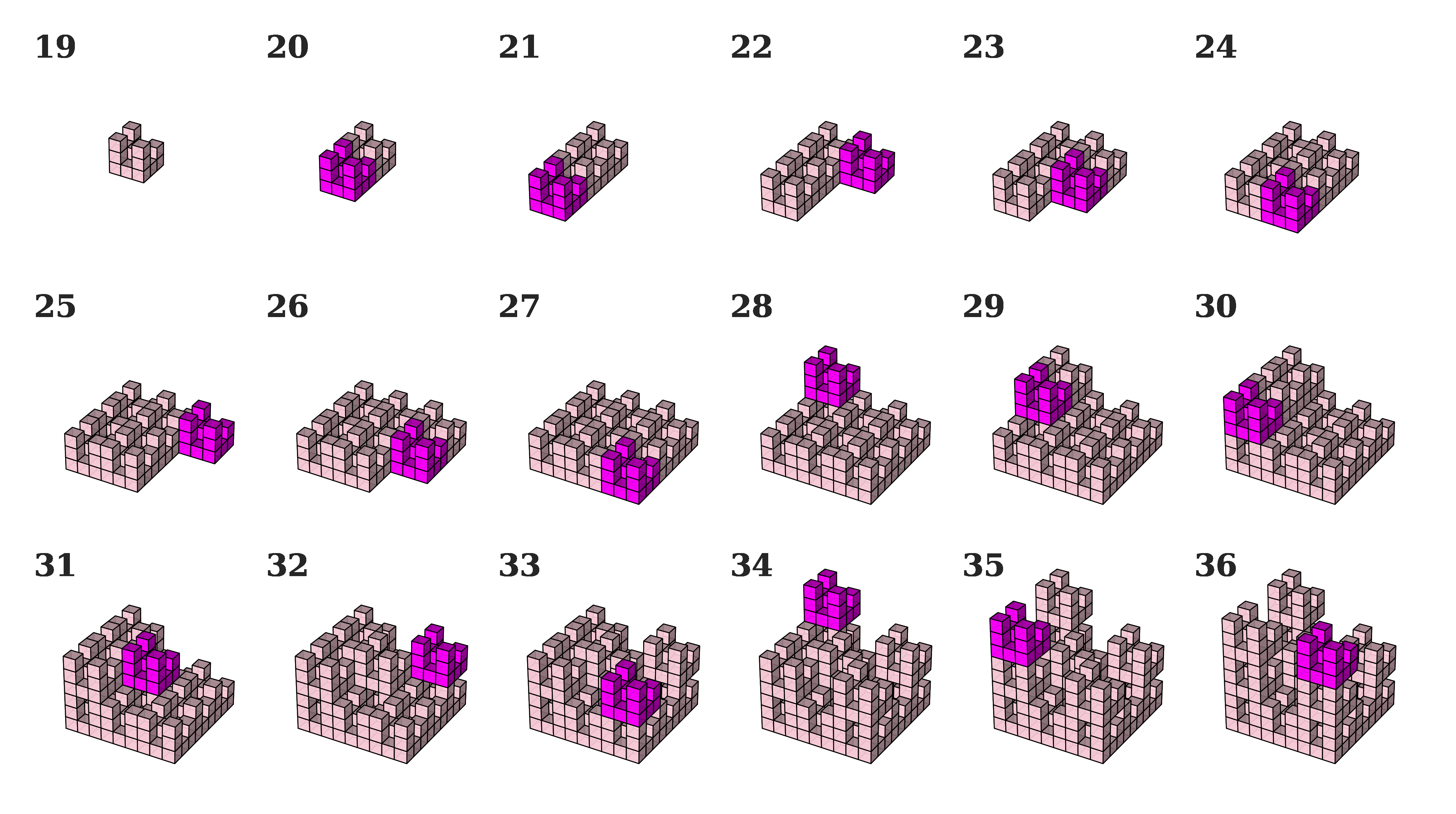}
    \caption{\textbf{Instructions for building the evolved single bladder fractal robot with the most scale invariant behavior.}
    The entire robot is a single actuator: it expands and contracts in unison according to a central pattern generator  (\href{https://youtu.be/FnHZBylM_JU}{video}).
    The basal robot (pictured in step 19) is designed in simulation by combining up to 27 voxels inside a 3-by-3-by-3 workspace (steps 1-18).
    The same steps are repeated using the basal robot instead of a single voxel (steps 19-36) to create a fractal assemblage.
    The physical robot fabricated according to these instructions can be seen, flipped right side up, in Fig.~\ref{fig:design2}.
    }
    \label{fig:instructions_single_actuator}
\end{figure*}